# An Iterative Feedback Mechanism for Improving Natural Language Class Descriptions in Open-Vocabulary Object Detection

Louis Y. Kim[a], Michelle Karker[a], Victoria Valledor[a], Seiyoung C. Lee[a],
Karl F. Brzoska[a], Margaret Duff[a], Anthony Palladino*[a]

[a]The Charles Stark Draper Laboratory, Inc., 555 Technology Square, Cambridge, MA 02139

## ABSTRACT

Recent advances in open-vocabulary object detection models will enable Automatic Target Recognition systems to be sustainable and repurposed by non-technical end-users for a variety of applications or missions. New, and potentially nuanced, classes can be defined with natural language text descriptions in the field, immediately before runtime, without needing to retrain the model. We present an approach for improving non-technical users' natural language text descriptions of their desired targets of interest, using a combination of analysis techniques on the text embeddings, and proper combinations of embeddings for contrastive examples. We quantify the improvement that our feedback mechanism provides by demonstrating performance with multiple publicly-available open-vocabulary object detection models.

**Keywords:** Computer vision, open-vocabulary object detection, human feedback, automatic target recognition

## 1. INTRODUCTION

Open-vocabulary object detection (OVOD) has the potential to revolutionize the field of automatic target recognition, by yielding more flexible, adaptable, and sustainable solutions. Instead of collecting labeled training datasets tailored to each specialized application and training custom object detection models, large vision-language models (VLMs) contain a wealth of semantic knowledge about the relationships between visual features and natural language text and perform well across domains with limited new input. In Draper's application-agnostic VLM-based automatic target recognition system [1], non-technical users can define the objects they wish to find, just before runtime, with simple text descriptions. In some cases, these descriptions may be too simple, contain ambiguities, or fail to contain sufficient differentiating features to prevent misclassification. We developed an approach that analyzes the text embeddings corresponding to the human-specified desired target, to identify potential lexical ambiguity or polysemy and guides the user to provide text descriptions with higher discriminative power. Improved descriptive inputs and fewer misclassifications eliminate the need for additional energy-intensive training or fine-tuning for the system to perform well on nuanced targets. The analysis leverages several techniques, including inter-class similarity assessments and concept decomposition (using sparse linear concept embeddings) to identify ambiguities. To avoid ambiguity, we allow the user to include contrastive descriptions, and we implement an extra step to account for these negative counter examples using custom embedding math. In this paper, we present a quantitative assessment of the improvement due to our feedback mechanism using multiple publicly-available pre-trained OVOD models, and images taken from standard OVOD benchmark datasets (OV-LVIS [2] and OV-COCO [3]). By incorporating this feedback mechanism, non-technical users' class descriptions will improve, thereby producing higher-performing, adaptable, and sustainable VLM-based ATR systems.

### 1.1 Related Work

This paper describes a tool to help non-technical end-users *use* AI, and we emphasize this as being distinct from a tool meant to help engineers train better ML models. There are a limited number of extant tools or techniques that are designed to improve performance of frozen VLMs at runtime (by the end user); for example, by adjusting class definitions/representations. Some of these techniques include prompt augmentation techniques, such as exploiting semantic relationships between concepts included in WordNet [4] to generate new coarse-grained or fine-grained prompts [5] or synonymous prompts [6]. Another prompt augmentation technique involves using LLMs as a knowledge base and producing detailed visual descriptions of a target class starting from just a class name [7, 8]. Further, a frozen VLM tasked with performing semantic segmentation could be improved at runtime by leveraging contrastive concepts generated by an LLM to improve single concept semantic segmentation [9].

On the other hand, there are a whole host of techniques intended to help researchers and engineers to better train their vision-language foundation models. Because they are implemented during the training stage, rather than being applied at runtime by a non-technical end-user, they are unrelated to the work presented in this paper. Nevertheless, we include them here for completeness. These techniques include: DVDet [10] which produces detailed and highly-related descriptors using region embeddings and large language model-based text alignment to augment the text at training time, OV-DQUO [11] introduces a training approach using denoising to improve classification of novel classes, AggDet [12] uses clustering centroids in visual space to identify region-text similarities, the authors of [13] introduced an evolutionary search algorithm using a large language model to learn visual attributes that maximally differentiate each class, meta-prompt learning [14] improves OVOD generalization to novel classes by leveraging the background proposals as negative cases of the foreground classes, the multi-modal attribute prompting (MAP) approach [15] uses visual and textual attribute prompting to improve semantic content and capture features in images, IntCoOp [16] performs attribute-level supervision during prompt tuning to improve generalization, and a part-stacked CNN [17] identifies important object parts for fine-grained recognition, perhaps leading to a stronger semantic understanding of the objects.

## 2. DEFINING TARGETS OF INTEREST

Unlike traditional object detection models that are trained on a fixed set of target classes, new open-vocabulary object detection models are no longer constrained to a fixed set of classes. Previously, engineers could do few-shot training or fine-tuning to add new classes and update the model, but now non-technical end-users are able to specify their own new target classes themselves without needing to change the model. Users can typically specify their desired classes in one of three ways: either using natural language text descriptions, or by providing one (or more) image exemplars, or using both natural language descriptions and image exemplars. A user is ultimately free to define their target class using as simple or as detailed of a description as they desired. This results in subjective, user-specific performance outcomes when applied in practice.

Automated approaches for generating more lengthy text descriptions starting from a simple class name have been demonstrated in MM-OVOD [18] which leveraged an LLM to generate text descriptions of the classes in their test set and found a modest 1.3% mAP improvement over the definitions using just the simple class names. On the contrary, the authors of VisionZip [19] showed that longer text (i.e., prompts) are better when it comes to LLMs, but not necessarily for VLMs. For VLMs, it seems that simpler/shorter class descriptions may result in better performance. EX2 [20] again used an LLM to generate descriptions, but their approach focused on including text that describes important visual features of the concepts.

For the feedback mechanism tool that we are presenting in this paper, we constrained our study to class definitions using text descriptions only. This likely aligns with the majority of real-world use cases, where it is difficult to quickly obtain an exemplar image of a novel class in the field. It is much easier for an end user to type the text description of the new class or tweak the details of the desired target, e.g., "they rolled down the windows on the target vehicle", so down weight the importance of the ATR-system finding tinted windows, rather than finding exemplar images of vehicles with their windows down.

The benefits to using natural language text to define and update class definitions are clear, as described above; however, there are also challenges. One challenge is the inherent multiplicity or polysemy in human language. Natural language spans the entire spectrum from being extremely vague to being very specific. For example, there are many visual interpretations of the word "house" (brick, stone, stucco, or clapboard siding, flat or pitched roof, various colors and sizes, etc.). We hypothesize that our feedback mechanism will demonstrate utility in alleviating unwanted multiplicity. Soldiers in the field may want a VLM-based ATR system to find all the tanks, and being so engulfed in their combat scenario, may not consider that there are other types of tanks which would be considered false alarms (water tank, propane tank, fish tank, tank top, etc.). In the following sections, we demonstrate how our feedback mechanism could help this end user improve their class definition to reduce multiplicity and isolate their target of interest.

## 3. FEEDBACK MECHANISMS

Our goal is to create a user-friendly system to help non-technical end users maximize the usefulness of emerging VLMs for ATR. To this end, we developed a tool that operates directly on the embedding vectors to improve their discriminatory power. It does not require any datasets and works with frozen models deployed in the final ATR capability.

VLMs historically have failed to properly handle negations; some researchers are attempting to build new VLMs that can understand negative concepts [21]. Our approach works because it does not rely on the model itself to understand the semantic queues in the natural language text indicating negation, instead operating directly on the embeddings. Our approach takes the embeddings that correspond to concepts a user has indicated as irrelevant and subtracts the average embedding from the original embedding with an adjustment factor, which we set to 0.3 for this initial study.

### 3.1 Leveraging Feedback using Class-specific Concept Embeddings

For the case where the user has just one target class, our tool provides feedback to the user, with guidance on how to adjust the initial class description. The tool allows the user to select the desired class, then analyzes encoded embedding to provide the user with context. The user can accept our system's feedback and provides adjustments to the class definition, such as adding positive or negative descriptions to differentiate the class of interest, or by subtracting concepts that are unrelated. The adjustments provided by the user are used to updates the class definition embedding. This can be used iteratively and when the user receives a satisfactory result, the tool can send the improved class definition to the edge platform (see Figure 1 below).

Our tool also leverages sparse linear concept embeddings (SpLiCE) [22], to associate the decomposition of CLIP embeddings into sets of related concepts with context, which our tool provides to the user as additional feedback. We anticipate that SpLiCE will aid in improving performance because VLMs sometimes relate generic attributes to concepts (e.g., a robin's main habitat is in North America), rather than visual descriptions of the concepts themselves (e.g., robins have dark grey or black feathers and a red breast), as demonstrated in SpLiCE, which lists the concepts. This may be why we sometimes see a drop in performance when un-selecting seemingly unrelated concepts; the VLM may find those concepts that the user found to be unrelated as important. This drop in performance is something that can be investigated in future work. Future work could also consist of using an approach similar to SpLiCE. This work would be to leverage a *k*-nearest-neighbors model to identify related embeddings [23].

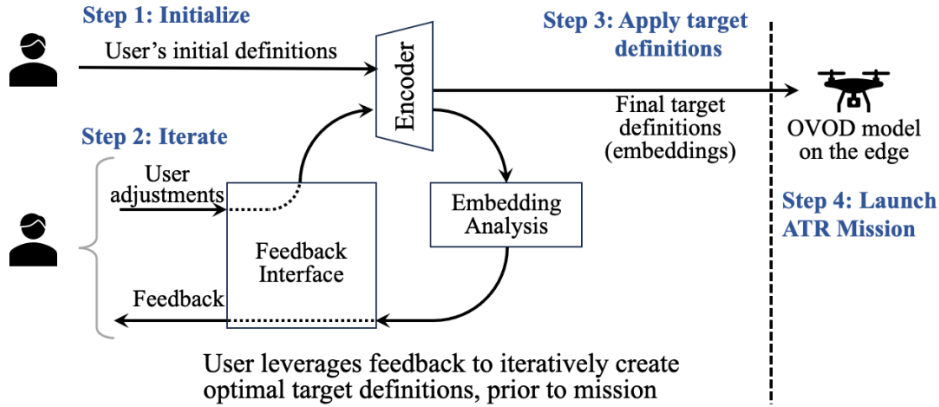

Figure 1. Our tool for refining target class definitions is used prior to mission, to achieve the best mission results. ATR is broadly defined here to include open-vocabulary object detection, and can include deployment on any platform, such as a drone, ground vehicle, satellite, or human body camera.

### 3.2 Leveraging Feedback using Inter-class Encoding Similarity

For the case where the user desires multiple targets simultaneously (e.g., both "passenger plane" and "military jet"), our tool computes the pairwise similarity between all pairs of target classes. We compute the cosine similarity, which can result in values from -1 (maximally dissimilar) to +1 (maximally similar). The feedback, in this case, is presented as a ranking of most similar and least similar classes. This suggests to the user that they should modify the text descriptions to help discriminate between a given class and the classes whose embeddings are most similar, by either adding embeddings of text describing visual features unique to a given class or subtracting embeddings of text that are common between those classes.

## 4. RESULTS

To evaluate the feedback mechanism, we designed open-vocabulary object detection experiments both with and without feedback. In these experiments, users respond to our tool's feedback to provide various adjustments to a fixed initial text description, "a jet plane", for existing open-vocabulary object detection models (MM-OVOD and YOLO-World) to better identify instances of fighter jets, compared to when the object detection models were provided with the initial text description input without any iterative adjustments.

In curating the dataset for the experiments, we incorporated distractor classes—airplanes and jet planes—that the object detection models might confuse with the target class, keeping in mind that in real-world application scenarios, there likely will be other objects distracting the target detection. From standard object detection benchmark datasets, COCO and LVIS, we selected 50 images containing airplanes, 15 images containing fighter jets, and 10 images containing jet planes. Since some images may contain multiple instances of an object class, this 75-image test set resulted in 94 instances of airplanes, 41 instances of fighter jets, and 22 instances of jet planes.

We measured object detection performance using mean average precision (mAP), with a key modification: we evaluated detections of the target class (fighter jets) across all images in the dataset, regardless of whether the target class was present. This adjustment accounts for potential false detections in images containing only distractor classes. Traditionally, object detection performance is measured by evaluating class detection within images that contain the specific class of interest, excluding those without it. Our modified evaluation method allows us to track changes in false detections within distractor-class images. We hypothesize that effective feedback will not only increase correct detections in images containing the target class but also reduce false detections in images containing distractor classes, leading to an overall higher detection performance.

For the purpose of proof-of-concept testing, six users conducted five different feedback experiments to enhance target class detection performance, consisting of adding and/or subtracting text, and toggling on/off related concepts. Figure 2 shows a partial screenshot of the experiment interface. Each experiment involved three iterations per user. The details of these experiments and their corresponding results are presented in the following subsections.

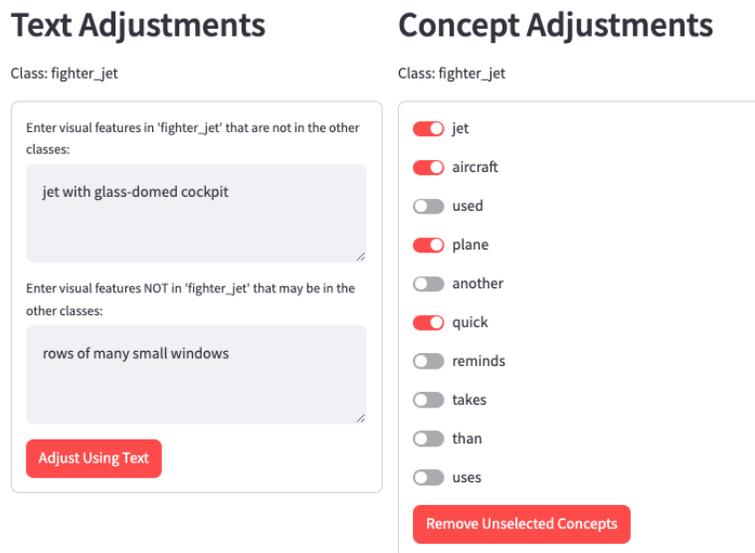

Figure 2. Screenshot showing a portion of the experiment interface that we developed to evaluate our feedback mechanism tool.

## 4.1 Experiment 1: Adding text description of distinguishing feature(s) of the target class

In this experiment, users utilized the feedback experiment interface to *add* text descriptions of distinguishing features of the target class, to help the object detection models better differentiate it from distractor classes. Each user performed three sequential feedback iterations, with the modified class definition embedding from each iteration provided to the object detection models to assess the impact of the feedback. Figure 3 presents the mean average precision (mAP) of the initial input compared to the users' average mAP across three iterations, with the standard error. While the maximum performance achieved through feedback exceeded the initial performance in all iterations, the average user performance declined by 0.7%, 3.4%, and 6.7% in iterations 1, 2, and 3 respectively. Table 1 summarizes the users' text inputs at each iteration and indicates whether the feedback resulted in the performance improvement or degradation.

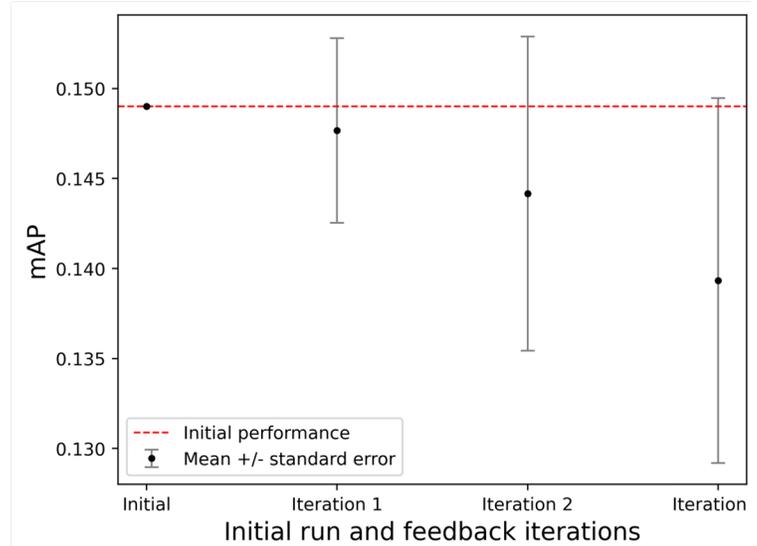

Figure 3. Experiment 1 detection performance results with the MM-OVOD model showing six users' average mAP with the standard error for each iteration, compared to mAP from the initial description.

Table 1. Text description added by each user at each iteration. Green texts indicate that the resulting detection performance was above the initial performance, red texts indicate that the resulting performance was below the initial performance, and black texts indicate no performance change.

|        | Iteration 1 | Iteration 2 | Iteration 3 |
|--------|-------------|-------------|-------------|
| User 1 | slim and sharp edges | sharp-pointed nose head | single dome-shaped cockpit |
| User 2 | jet with glass-domed cockpit | fighter jet with two vertical tail fins | grey or silver attack aircraft |
| User 3 | fighter jet | military airplane | warplane |
| User 4 | sleek, high-speed military plane | gray or camouflage with a matte surface | tinted glass canopy |
| User 5 | military aircraft with weapons | gray or black jet with streamlined fuselage | combat aircraft for tactical missions |
| User 6 | small trapezoidal wings | includes bombs, missiles, or other firepower | highly maneuverable in the air |

## 4.2 Experiment 2: Removing text description of non-relevant feature(s) of the target class

In this experiment, users utilized the experiment interface to *remove* text descriptions of irrelevant features of the target class, allowing the object detection models to better distinguish it from distractor classes. Unlike in Experiment 1, users specified features that are absent in the target class but potentially present in the distractor classes, which could otherwise lead to misclassification. **Error! Reference source not found.** presents the mean average precision (mAP) of the initial input compared to the users' average mAP across three iterations, along with the standard error. While the performance varied among users due to the subjective nature of text descriptions provided during feedback, the average performances improved by 0.7% and 3.6% in iterations 1 and 2, respectively, but declined by 8.7% in iteration 3.

Table 2 summarizes the users' text inputs at each iteration and indicates whether the feedback resulted in performance improvement or degradation.

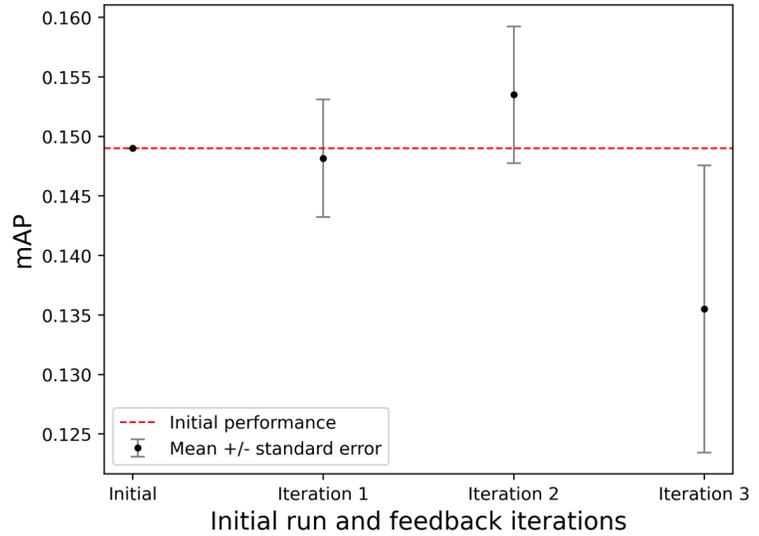

Figure 4. Experiment 2 detection performance results with the MM-OVOD model showing six users' average mAP along with the standard error for each iteration, compared to mAP from the initial description.

Table 2. Text description of each user at each iteration. Green texts indicate that the resulting detection performance was above the initial performance, and red texts indicate that the resulting performance was below the initial performance.

|        | Iteration 1 | Iteration 2 | Iteration 3 |
|--------|-------------|-------------|-------------|
| User 1 | passenger windows | tube-shaped body | propeller |
| User 2 | rows of many small windows | large white passenger plane | airline company logo |
| User 3 | white | turbine engine | commercial jet |
| User 4 | large under-wing jet engines | side door and open cabin | boxy fuselage with rear cargo ramp |
| User 5 | round fuselage | many little windows | white plane carrying passengers |
| User 6 | long thin wings | high passenger capacity | cylindrical engines under the wings |

## 4.3 Experiment 3: Simultaneously adding text description of distinguishing feature(s) of the target class and removing text description of non-relevant feature(s) of the target class

In this experiment, users simultaneously added descriptions of distinguishing features and removed descriptions of irrelevant features by inputting the same text modifications from the previous two experiments.

Figure 5 presents the mean average precision (mAP) of the initial input compared to the users' average mAP across three iterations, along with the standard error. The average performance improved by 2.0% and 2.7% in iterations 1 and 2, respectively, but declined by 3.4% in iteration 3.

Table 3 summarizes the users' text inputs at each iteration and indicates whether the feedback resulted in the performance improvement or degradation.

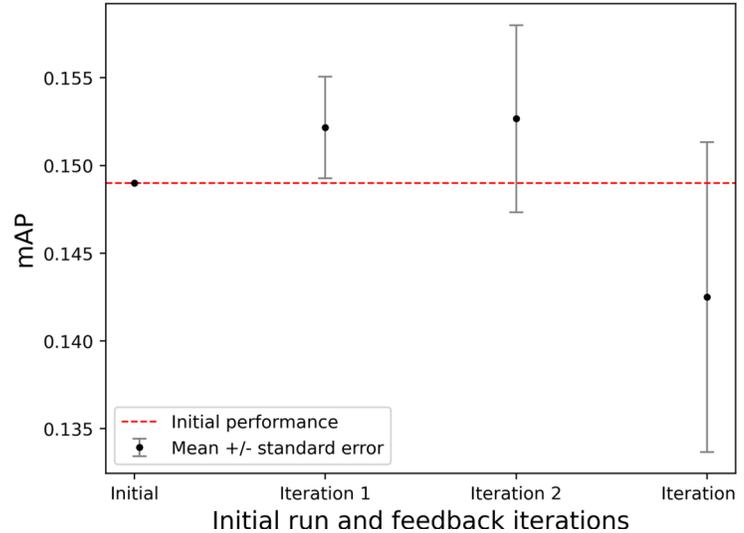

Figure 5. Experiment 3 detection performance results with the MM-OVOD model showing six users' average mAP along with the standard error for each iteration, compared to mAP from the initial description.

Table 3. Text description of each user at each iteration. Green texts indicate that the resulting detection performance was above the initial performance, and red texts indicate that the resulting performance was below the initial performance.

|  |  | Iteration 1 | Iteration 2 | Iteration 3 |
|---|---|---|---|---|
| User 1 | Added | slim and sharp edges | sharp-pointed nose head | single dome-shaped cockpit |
|  | Removed | passenger windows | tube-shaped body | propeller |
| User 2 | Added | jet with glass-domed cockpit | fighter jet with two vertical tail fins | grey or silver attack aircraft |
|  | Removed | rows of many small windows | large white passenger plane | airline company logo |
| User 3 | Added | fighter jet | military airplane | warplane |
|  | Removed | white | turbine engine | commercial jet |
| User 4 | Added | sleek, high-speed military plane | gray or camouflage with matte surface | tinted glass canopy |
|  | Removed | large under-wing jet engines | side doors and open cabin | boxy fuselage with rear cargo ramp |
| User 5 | Added | military aircraft with weapons | gray or black jet with streamlined fuselage | combat aircraft for tactical missions |
|  | Removed | round fuselage | many little windows | white plane carrying passengers |
| User 6 | Added | small trapezoidal wings | includes bombs, missiles, or other firepower | highly maneuverable in the air |
|  | Removed | long thin wings | high passenger capacity | cylindrical engines under the wings |

## 4.4 Experiment 4: Removing non-relevant concepts

In this experiment, users were presented with a list of the 10 most relevant concepts identified by the SpLiCE method in the feedback interface. At each iteration, they unselected one or more concepts they deemed irrelevant to the target class.

Figure 6 compares the mean average precision (mAP) of the initial input to the users' average mAP across three iterations. Since all users unselected the same set of irrelevant concepts in each iteration, the detection performance remained consistent across users. Given that the user's produced the exact same results, there is no standard error. Performance improved by 12.8%, 17.4%, and 16.8% in iterations 1, 2, and 3, respectively.

Table 4 summarizes the concepts identified by the SpLiCE method which were unselected by users as being irrelevant to the target class.

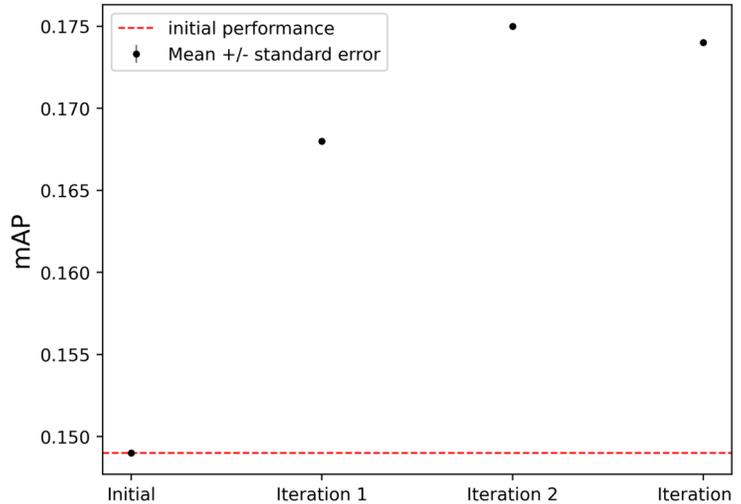

Figure 6. Experiment 4 detection performance results with the MM-OVOD model showing six users' average mAP for each iteration, compared to mAP from the initial description. The error bars are not present here since all users selected the same adjustments in all iterations, which resulted in identical performance.

Table 4. Selected and unselected concepts at each iteration. Note that all users made the same selections.

|  | **Iteration 1** | **Iteration 2** | **Iteration 3** |
|---|---|---|---|
| **Unselected concepts** | another, seen, than, reminds, potential, used, being something | another, seen, uses, takes, than, reminds | joke, slightly, arts, floppy, lithograph |
| **Selected concepts** | jet, aircraft | jet, aircraft, quick, plane | jet, aircraft, plane, quick, jets |

## 4.5 Experiment 5: Adding and removing textual features and removing non-relevant concepts

In this experiment, users simultaneously added descriptions of distinguishing features and removed descriptions of irrelevant features, similar to Experiment 3. In addition, after providing the text inputs, users were presented with a list of concepts from the SpLiCE method, and asked to unselect non-relevant concepts, following the approach used in Experiment 4.

Figure 7 compares the mean average precision (mAP) of the initial embedding to the users' average mAP across three iterations, along with the standard error. The average performance improved by 13.4%, 14.8%, and 13.4% for iterations 1, 2, and 3, respectively.

The same text inputs from Table 3 were used in this experiment, while Table 5 summarizes concepts each user unselected at each iteration.

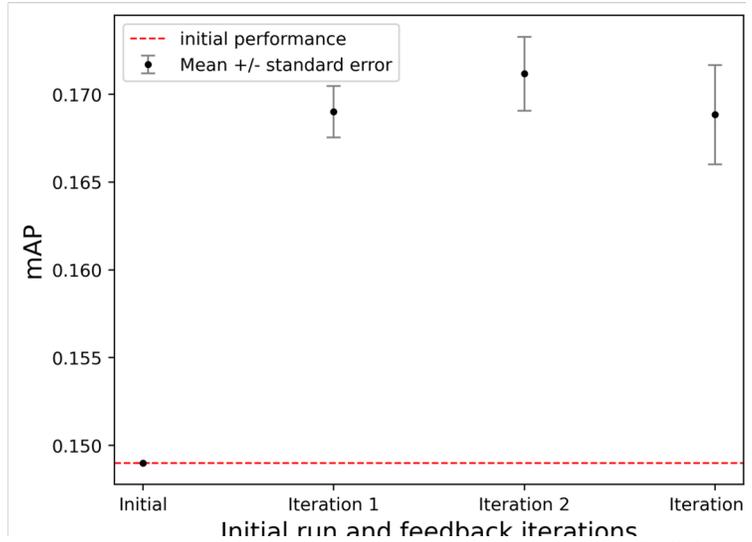

Figure 7. Experiment 5 detection performance results with the MM-OVOD model showing six users' average mAP along with the standard error for each iteration, compared to mAP from the initial description.

Table 5. Text description added by each user at each iteration.

|  |  | Iteration 1 | Iteration 2 | Iteration 3 |
|---|---|---|---|---|
| User 1 | Unselected concepts | another, reminds, slightly, than, being, attempt, uses, takes | another, slightly, reminds, job, attempt, takes, being | pictured, cheap, writes, described, hurt, scam |
|  | Selected concepts | jet, aircraft | jet, aircraft, sharp | Jet, aircraft, shape, sleek |
| User 2 | Unselected concepts | used, another, reminds, takes, than, uses | that, something, science, uses, dick, moto | concept, supercar, dick, tank, shark |
|  | Selected concepts | jet, aircraft, plane, quick | jet, aircraft, designed, quick | jet, aircraft, jets, sleek, fighter |
| User 3 | Unselected concepts | another, seen, reminds, than, potential, something, takes, used | athlete, knife, motorcycle, floppy, lithograph | dolphins, yacht, baggage, dragonfly, submarine |
|  | Selected concepts | jet, aircraft | jet, aircraft, plane, jets, bullet | jet, aircraft, jets, plane, aviator |
| User 4 | Unselected concepts | reminds, another, used, attempt, potential, takes | reminds, attempt, slightly, alt, subtle | reminds, testing, subtle, surface |
|  | Selected concepts | jet, aircraft, plane, quick | jet, plane, military, aircraft, quick | jet, plane, military, sleek, quick, presence |
| User 5 | Unselected concepts | another, than, potential, seen, something, attempt, reminds | but, uses, another, speaks, takes | supercar, arts, speaks |
|  | Selected concepts | jet, aircraft, military | jet, aircraft, military, quick, terrorism | jet, aircraft, military, jets, sleek, terrorism, tactical |
| User 6 | Unselected concepts | another, than, seen, used, potential, uses, styled | another, uses, than, takes, arts, several, something | another, subtle, its, failure, uses |
|  | Selected concepts | jet, aircraft, quick | jet, aircraft, quick | jet, aircraft, quick, plan, bold |

## 4.6 Summary of Results

Table 6 summarizes our overall results with the MM-OVOD model. While not all feedback iterations led to improved detection performance, the average performance across all experiments showed an increase of 2.3% to 7.0%. Experiment 4 demonstrated the most substantial improvement, with performance gains ranging from 12.8% to 17.4% compared to the initial description. Notably, in Experiment 4, users took as little as 5 to 10 seconds per feedback iteration to unselect non-relevant concepts before running the object detection model. The fact that a few seconds of user intervention—without additional model training or fine-tuning—led to significant performance improvement highlights the potential of the proposed feedback mechanism for enhancing open-vocabulary object detection in real world applications.

In contrast, Experiments 1, 2, and 3, which relied solely on textual adjustments, yielded mixed results. The high variation in performance can be attributed in part to the subjective nature of free-form text inputs, as different users may have differing opinions on what constitutes important or non-relevant features. Additionally, adding a seemingly reasonable distinguishing feature such as *"sharp-pointed nose head"* (since jet fighter have sharper noses compared to commercial airliners or jet planes) and removing an intuitive non-relevant feature such as *"airline company logo"* (which commercial airliners have but jet fighter dos not) both unexpectedly led to performance degradation. Several factors may have contributed to these counterintuitive results. First, the 512-dimensional CLIP embeddings provide a numerical estimation of textual inputs, which may not fully capture their exact meaning. Consequently, simple embedding addition or subtraction may not precisely modify the intended feature representation, leading to outcomes different from what users expect. Further, open-vocabulary object detection models have inherent limitations. While these models can detect objects unseen during training or with limited training data, they often suffer from low detection accuracy and high sensitivity to text prompt quality. As a result, detecting less frequently studied classes, such as jet fighters, may be particularly sensitive to variation in descriptive feedback. Note that YOLO-World was unable to make any correct detections in our set up with the distractors, and the feedback experiments were not helpful in the case where the underlying OVOD model did not work.

Regarding the number of feedback iterations, the initial expectation was that refining descriptions over multiple iterations would lead to continuous detection improvements; however, results from these experiments indicate that additional iterations do not necessarily enhance detection performance. Several factors, such as the subjectivity of user feedback, error in embedding representation, and the inherent sensitivity of detection models to open text descriptions, may have contributed to the non-monotonic trend in performance. Further, it was observed that non-relevant concepts removed in one iteration, such as *"another"* and *"seen"*, reappeared in subsequent iterations, prompting users to remove them again. Repeatedly eliminating the same embedding vector may have unintentionally weakened the original embedding representation, leading to unexpected modifications and potential degradation in performance.

Table 6. Quantification of the improvement in mean average precision (mAP) for the best experiment and averaged over all experiments.

|  | mAP | | | |
| --- | --- | --- | --- | --- |
| **Experiment** | **Baseline** | **1 Iteration** | **2 Iterations** | **3 Iterations** |
| **Avg. All Experiments** | 0.149 | 0.157 (5.6% ↑) | 0.160 (7.0% ↑) | 0.152 (2.3% ↑) |
| **Experiment 4** | 0.149 | 0.168 (12.8% ↑) | 0.175 (17.4% ↑) | 0.174 (16.8% ↑) |

## 5. FUTURE WORK

While our experimental results demonstrated the promising potential of the proposed feedback mechanism, further research is needed to evaluate its generalizability and robustness. Future experiments should incorporate a wider range of object categories to assess whether the method performs consistently across different classes. Additionally, increasing the number of users will be necessary to more accurately evaluate variability in performance as a result of individual differences in interpretation.

In this study, we developed a novel feedback mechanism and explored simple embedding adjustments by adding embeddings of distinguishing features, subtracting embeddings of non-relevant features, and removing embeddings of irrelevant SpLiCE concepts. We made these modifications using a few arbitrarily chosen weights for embedding adjustments. A potential direction for future work is to develop models that optimize these weights to maximize detection performance. For instance, if an embedding lies within a dense cluster, it may indicate ambiguity or multiplicity, increasing the likelihood of false detections for visually similar but distinct classes. Computing optimal weights for embedding adjustments could help shift the embedding representation away from such clusters, improving the model's ability to distinguish the target class from distractors.

To address the main limitation of subjective and potentially unhelpful textual feedback, recent advancements such as CLIP-InterpreT [24, 25] can be leveraged. CLIP-InterpreT enhances interpretability in CLIP models by retrieving nearest-neighbor images for a given input text using different attention heads. Integrating CLIP-InterpreT into our feedback tool would allow users to visualize how the vision-language model (VLM) interprets their initial descriptions. By providing visual guidance, users could refine their textual feedback more effectively, leading to more precise and useful adjustments for detecting the target object.

## 6. CONCLUSIONS

We proposed a novel feedback mechanism that enables non-technical end-users to improve open-vocabulary object detection performance through efficient feedback mechanisms, at runtime and without changing the underlying model. With just a few mouse clicks and/or simple text input at runtime to update the embedding of the initial class definition, we demonstrated detection performance improvements of up to 17.6% in mean average precision (mAP) using existing popular open-vocabulary object detection models.

The proposed solution will enable non-technical end-users to leverage the rapidly expanding library of open-vocabulary object detection models for enhanced automated target detection across various applications. Additionally, the future work outlined in this work will further refine and expand the capabilities of this feedback mechanism, improving its effectiveness and applicability.

## ACKNOWLEDGEMENTS

The authors would like to thank the SPIE 2025 Automatic Target Recognition XXXV conference organizers and the anonymous referees for their valuable comments. We would like to thank The Charles Stark Draper Laboratory, Inc. for providing the Internal Research & Development funding that supported this work.